\documentclass[10pt,twocolumn,letterpaper]{article}

\usepackage{iccv}
\usepackage{times}
\usepackage{epsfig}
\usepackage{graphicx}
\usepackage{amsmath}
\usepackage{amssymb}

\usepackage{url}
\usepackage{caption,subcaption}
\usepackage{multirow}
\usepackage{arydshln}
\usepackage{appendix}

\usepackage[pagebackref=true,breaklinks=true,letterpaper=true,colorlinks,bookmarks=false]{hyperref}

\DeclareMathOperator*{\argmin}{arg\,min}

\iccvfinalcopy 

\def\iccvPaperID{****} 

\ificcvfinal\fi

\begin{document}

\title{Gradient-Coherent Strong Regularization for \\ Deep Neural Networks with Stochastic Gradient Descent}


\author{Dae Hoon Park\textsuperscript{1}, 
  Chiu Man Ho\thanks{Most work was done while at Huawei Research America.}\textsuperscript{~~2},
  Yi Chang$^*$\textsuperscript{3}, 
  Huaqing Zhang\textsuperscript{1}\\
  \textsuperscript{1}Huawei Research America, 
  \textsuperscript{2}Innopeak Technology, 
  \textsuperscript{3}Jilin University\\
  daehpark@gmail.com,
  chiuman100@gmail.com,
  yichang@acm.org,
  markasjunior@gmail.com
}

\maketitle

\begin{abstract}
Regularization plays an important role in generalization of deep neural networks, which are often prone to overfitting with their numerous parameters.
L1 and L2 regularizers are common regularization tools in machine learning with their simplicity and effectiveness.
However, we observe that imposing strong L1 or L2 regularization with stochastic gradient descent on deep neural networks easily fails, which limits the generalization ability of the underlying neural networks.
To understand this phenomenon, we first investigate how and why learning fails when strong regularization is imposed on deep neural networks.
We then propose a novel method, gradient-coherent strong regularization, which imposes regularization only when the gradients are kept coherent in the presence of strong regularization.
Experiments are performed with multiple deep architectures on three benchmark data sets for image recognition.
Experimental results show that our proposed approach indeed endures strong regularization and significantly improves both accuracy and compression (up to 9.9x), which could not be achieved otherwise.

\end{abstract}

\section{Introduction} \label{sec:introduction}
Regularization is a common tool for machine learning to prevent overfitting.
Deep neural networks (DNNs), which have shown huge success in many areas such as computer vision \cite{krizhevsky2012imagenet,simonyan2014very,he2016deep} and speech recognition \cite{hinton2012deep}, often contain a number of trainable parameters in multiple layers with non-linear activation functions, in order to gain enough expressive power.
However, DNNs with such many parameters are often prone to overfitting, so the need for regularization has been emphasized.
While regularization techniques such as dropout \cite{srivastava2014dropout} have been proposed to solve the problem, the traditional L1 and L2 regularizers have cooperated well with them to further improve the performance significantly.
For example, our empirical results on image recognition show that strong L2 regularization to DNNs that already has dropout layers can reduce the error rate by up to 24\% on a benchmark data set.
However, we observe that DNNs easily fail to learn when strong L1 or L2 regularization is imposed (by a large value of the regularization parameter) with stochastic gradient descent.

Strong regularization on DNNs is often desired for two main reasons.
First, strong regularization can result in better generalization especially when a model is greatly overfitted, which is often the case for DNNs.
Strong regularization yields a simple solution, which is less prone to overfitting and preferred over complex ones with the principle of Occam's razor.
Second, strong regularization by sparse regularizers such as L1 regularizer compresses a solution into a sparse one while keeping or even improving its generalization.
As DNNs typically consist of numerous parameters, such sparse solutions in sparse matrices may reduce a storage overhead or reside in a memory with less energy consumption.
For example, a 9x compressed competitive DNN solution in sparse matrices achieved a storage overhead of only about 16\% of non-compressed one while it resides in on-chip SRAM instead of off-chip DRAM that consumes more than 100x energy \cite{han2015learning}.

Unfortunately, imposing strong L1 or L2 regularization on DNNs is difficult with stochastic gradient descent.
Indeed, difficulties related to non-convexity and the use of stochastic gradient descent are often overlooked in the literature.
We observe and analyze how and why learning fails in DNNs with strong L1 or L2 regularization.
We hypothesize that the dominance of gradients from the regularization term, caused by strong regularization, iteratively diminishes magnitudes of both weights and gradients.
To prevent the failure in learning, we propose a novel approach ``gradient-coherent strong regularization'' that imposes strong regularization only when the gradients from regularization do not obstruct learning.
That is, if the sum of regularization gradients and loss gradients is not coherent with the loss gradients, our approach does not impose regularization.
Experiments were performed for traditional DNNs and data sets, and the results indicate that our proposed approach indeed achieves strong regularization, resulting in both better generalization and more compression.
Our main contributions in this work can be summarized as:
\begin{itemize}
    \item We provide the first novel analysis how and why learning fails with \textit{strong} L1/L2 regularization in DNNs. To the best of our knowledge, there is no existing work that theoretically or empirically analyzes this phenomenon.
    \item We propose a novel approach, gradient-coherent strong regularization, that selectively imposes strong L1/L2 regularization to avoid failure in learning.
    \item We perform experiments with multiple deep architectures on three benchmark data sets for image recognition. Our proposed approach does not fail for strong regularization and significantly improves the accuracy. It also compresses DNNs up to 9.9x without losing its accuracy.
\end{itemize}

\section{Problem Analysis}\label{sec:problem}

\subsection{DNNs and Strong L1/L2 Regularization}
Let us denote a generic DNN by $\mathbf{\hat{y}} = f(\mathbf{x};\mathbf{w})$ where $\mathbf{x} \in \mathbb{R}^d$ is an input vector, $\mathbf{w} \in \mathbb{R}^n$ is a flattened vector of all parameters in the network $f$, and $\mathbf{\hat{y}} \in \mathbb{R}^c$ is an output vector after feed-forwarding $\mathbf{x}$ through multiple layers in $f$.
The network $f$ is trained by finding an optimal set of $\mathbf{w}$ with the following objective function.
\begin{equation}\label{eq:objective}
	\begin{aligned}
	\mathbf{w^*} = \argmin_{\mathbf{w}} \frac{1}{|\mathcal{D}|} \sum_{(\mathbf{x,y}) \in \mathcal{D}} \mathcal{L}(f(\mathbf{x};\mathbf{w}), \mathbf{y}) + \lambda \Omega(\mathbf{w})
	 \end{aligned}
\end{equation}
where $\mathcal{D}$ is the training data, and $\mathcal{L}$ is the loss function, which is usually cross-entropy loss for classification tasks.
Here, the regularization term $\lambda\Omega(\mathbf{w})$ is added to impose penalty on complexity of the solution, and $\lambda$, which we refer to as \textit{regularization strength}, is set to zero for non-regularized models.
A higher value of $\lambda$ thus means stronger regularization.

With a gradient descent method, each model parameter at time $t$, $w^{(t)}$, is updated with the following formula:
\begin{equation}\label{eq:gd}
\small
	\begin{aligned}
	w^{(t+1)} &= w^{(t)} - \alpha \left.\left(\,
	\frac{\partial \mathcal{L}}{\partial w} + 
	\lambda \frac{\partial \Omega }{\partial w}
	\,\right)\right|_{w=w^{(t)}} \\
    \frac{\partial \Omega}{\partial w} &=
    \begin{cases}
    2w^{(t)}, & \text{if } \Omega(\mathbf{w}) = ||\mathbf{w}||^2_2\\
    \text{sign}(w^{(t)}), & \text{if } \Omega(\mathbf{w}) = ||\mathbf{w}||_1 \text{ and } w^{(t)} \neq 0
    \end{cases}
    \end{aligned}
\normalsize
\end{equation}
where $\alpha$ is a learning rate.
We refer to $\frac{\partial \mathcal{L}}{\partial w}$ and $\frac{\partial \Omega }{\partial w}$ as $\nabla \mathcal{L}$ and $\nabla \Omega$ throughout the paper.
The regularization function $\Omega(\mathbf{w}) = ||\mathbf{w}||^2_2$ is an L2 regularizer, which is the most commonly used regularizer and is also called as \textit{weight decay} in deep learning literature.
As shown, L2 regularizer reduces the magnitude of a parameter proportionally to it.
It thus imposes a greater penalty on parameters with greater magnitudes and a less penalty on parameters with less magnitudes, yielding a simple and effective solution that is less prone to overfitting.
On the other hand, L1 regularizer (also known as Lasso \cite{tibshirani1996regression}): $\Omega(\mathbf{w}) = ||\mathbf{w}||_1$ is often employed to induce sparsity in the solution (\textit{i.e.,} make a portion of $\mathbf{w}$ zero), by imposing the same magnitude ($\lambda$) of penalty on all parameters.
In both L1 and L2 regularizers, \textit{strong regularization} thus means great penalty on  magnitudes of the parameters.

\subsection{Imposing Strong Regularization Makes Learning Fail.} \label{sec:fail}

\begin{figure}[hbt] 
\centering
\begin{subfigure}{.5\columnwidth}
    \centering
    \includegraphics[width=0.9\linewidth]{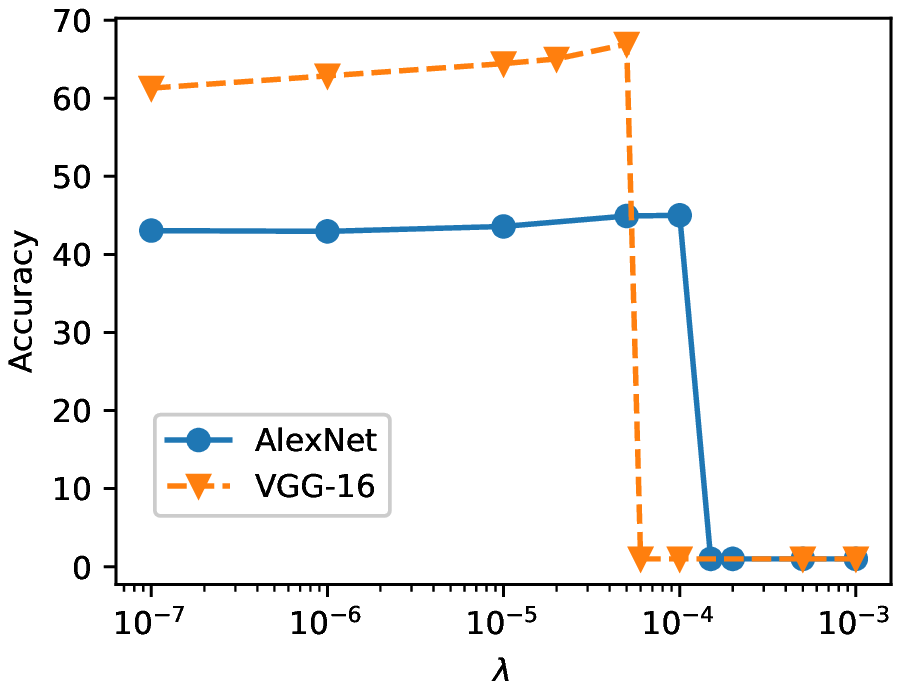}
    \caption{L1 regularization}
    \label{fig:accuracy_drop_1}
\end{subfigure}%
\begin{subfigure}{.5\columnwidth}
    \centering
    \includegraphics[width=0.9\linewidth]{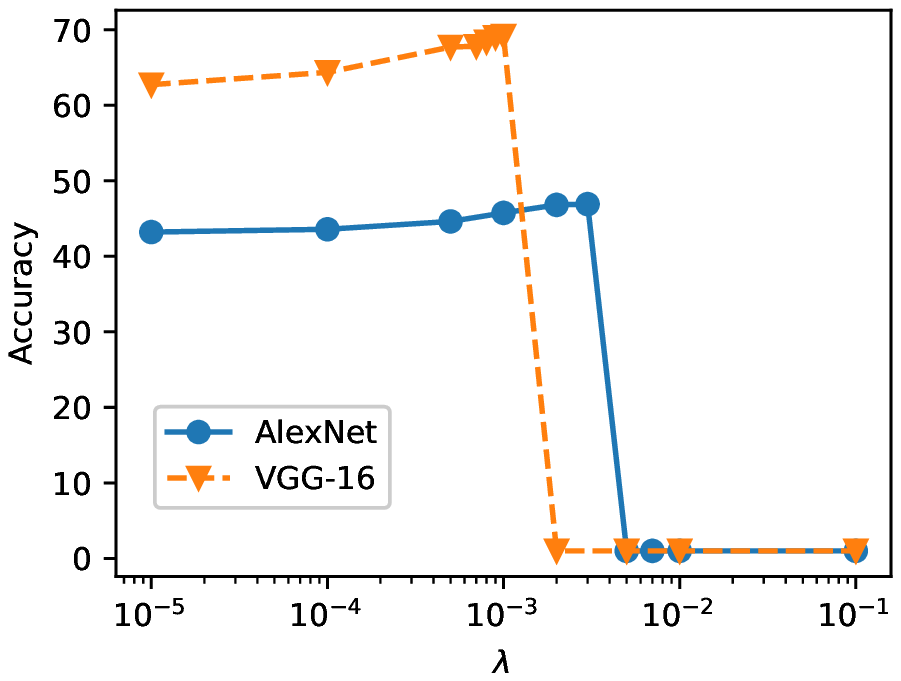}
    \caption{L2 regularization}
    \label{fig:accuracy_drop_2}
\end{subfigure}
\caption{ Validation accuracies on CIFAR-100. Note the sharp accuracy drop.  }
\label{fig:training_fails}
\end{figure}

\begin{figure*}[tb] 
\centering
\begin{subfigure}{.2\textwidth}
    \centering
    \includegraphics[width=1.0\linewidth]{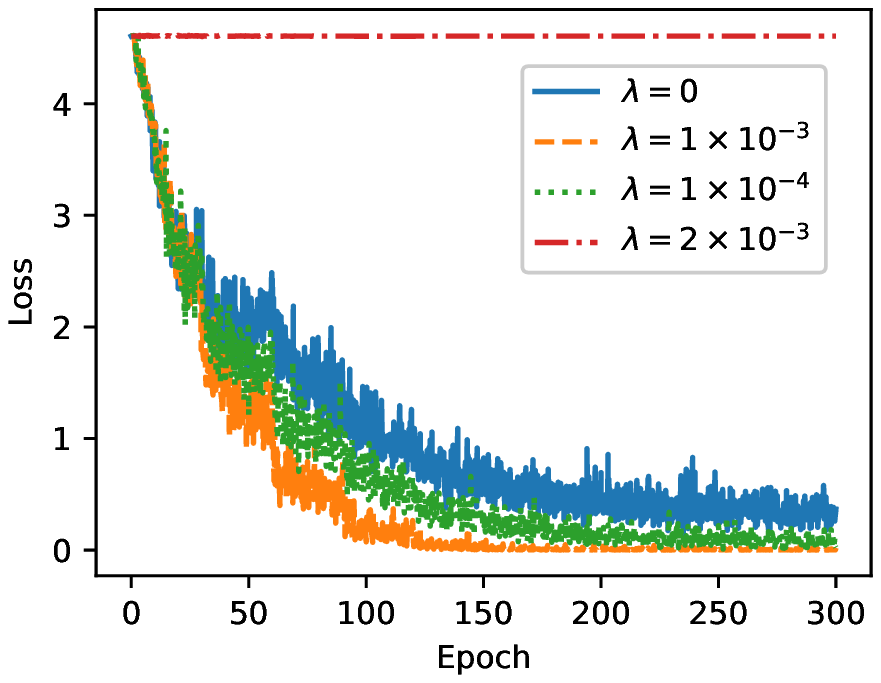}
    \caption{Training loss (L2 Reg.)}
    \label{fig:training_loss}
\end{subfigure}%
\begin{subfigure}{.2\textwidth}
    \centering
    \includegraphics[width=1.0\linewidth]{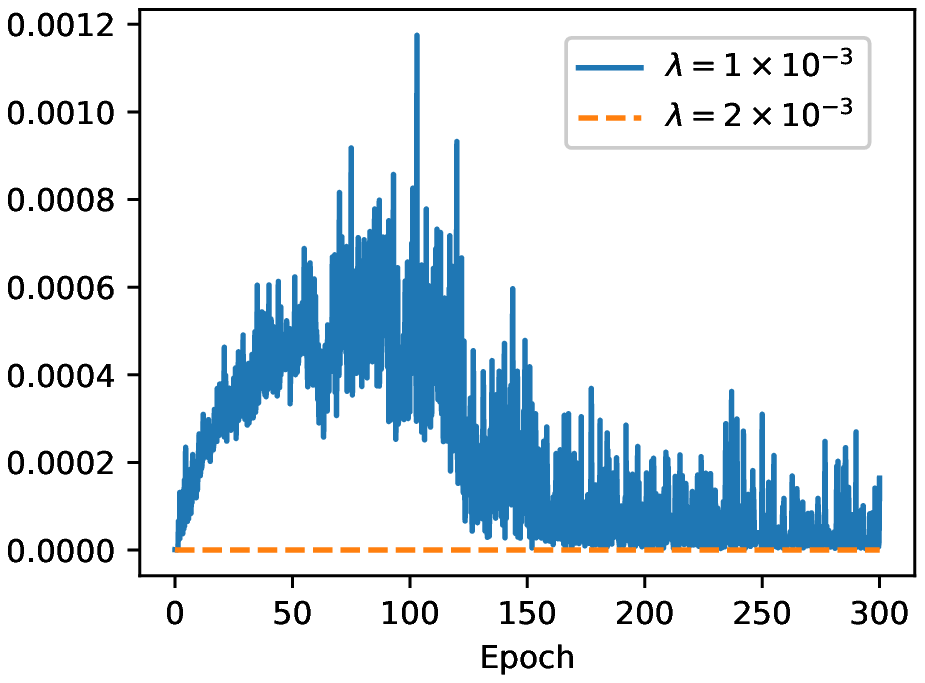}
    \caption{Avg. $|\nabla \mathcal{L}|$ (L2 Reg.)}
    \label{fig:avg_abs_grad}
\end{subfigure}%
\begin{subfigure}{.2\textwidth}
    \centering
    \includegraphics[width=1.0\linewidth]{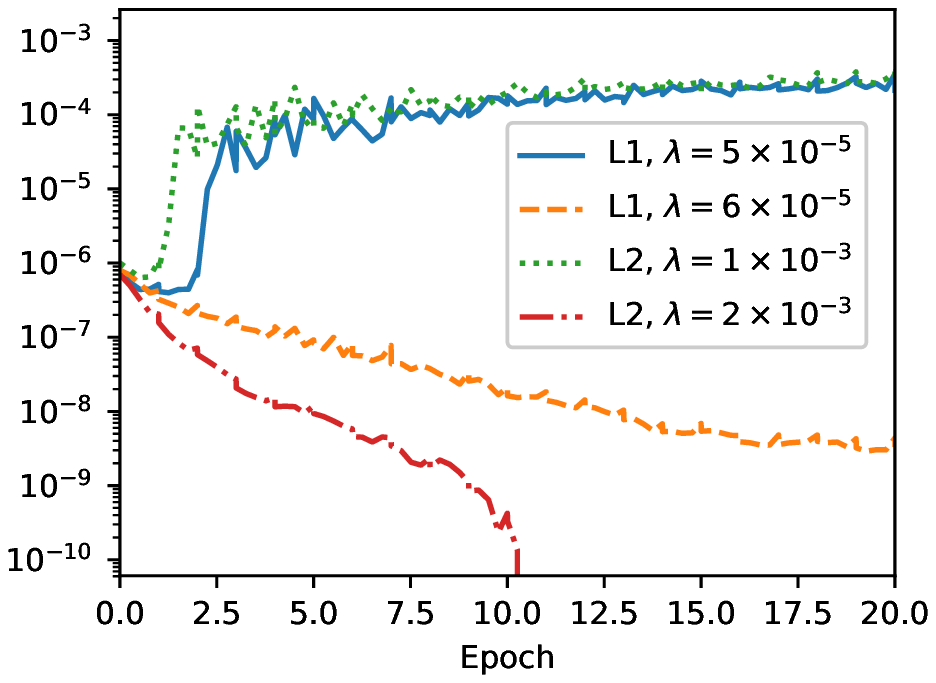}
    \caption{Avg. $|\nabla \mathcal{L}|$ (close-up) }
    \label{fig:avg_abs_grad_log}
\end{subfigure}%
\begin{subfigure}{.2\textwidth}
    \centering
    \includegraphics[width=1.0\linewidth]{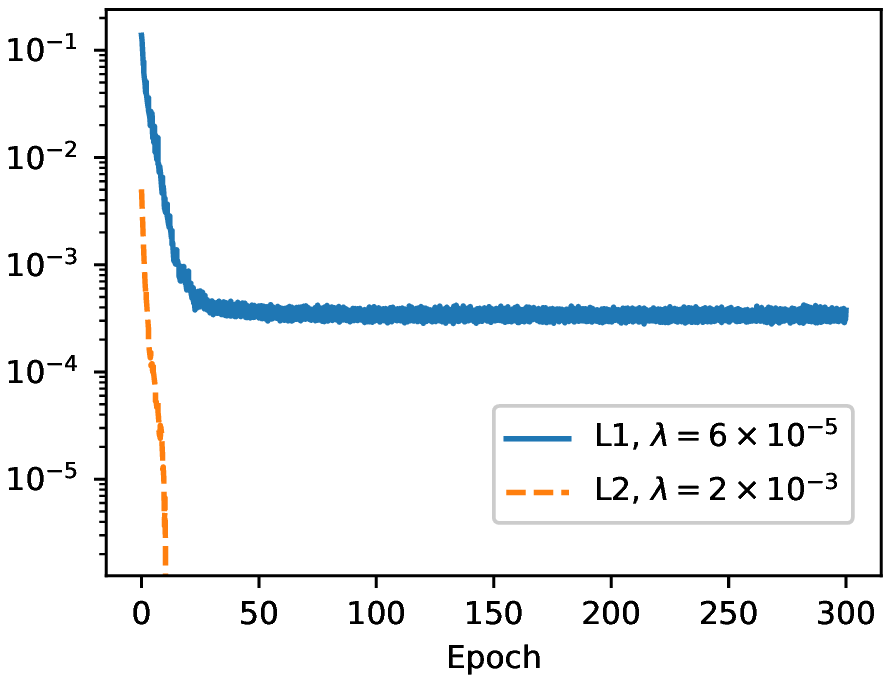} 
    \caption{Avg. $\frac{ |\nabla \mathcal{L}|}{ |\nabla \mathcal{L}| + \lambda|\nabla \Omega|}$} 
    \label{fig:grad_portion}
\end{subfigure}%
\begin{subfigure}{.2\textwidth}
    \centering
    \includegraphics[width=1.0\linewidth]{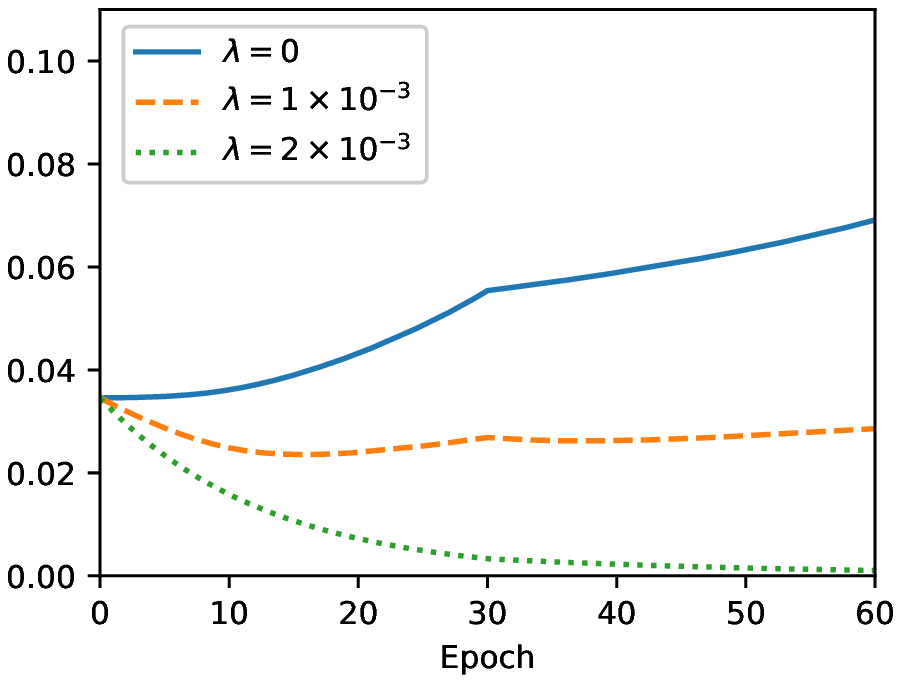} 
    \caption{Avg. $|w|$ (L2 Reg.)} 
    \label{fig:weight_comp}
\end{subfigure}
\caption{Statistics for different $\lambda$ by VGG-16 on CIFAR-100. Best shown in color.}
\label{fig:gradients}
\end{figure*}

Strong regularization is especially useful for deep learning because the DNNs often contain a large number of parameters while the training data are relatively limited in practice.
However, we observe a phenomenon where learning suddenly fails when strong regularization is imposed with stochastic gradient descent, which is the most commonly used solver for deep learning.
The example of the phenomenon is depicted in Figure \ref{fig:training_fails}.
The architectures VGG-16 \cite{simonyan2014very} and AlexNet \cite{krizhevsky2012imagenet} were employed for the dataset CIFAR-100 \cite{krizhevsky2009learning}.\footnote{Details of the experiment setting are described in the experiments section.}
As shown in this example, the accuracy increases as we enforce stronger regularization with greater $\lambda$.
However, it suddenly drops to 1.0\% after enforcing a little stronger regularization, which means that the model fails to learn.\footnote{The sudden drop still exists on a linear scale of $\lambda$.}
This observation raises three questions: (\texttt{i}) How and why does learning fail with strong regularization in deep neural networks? (\texttt{ii})  How can we avoid the failure? (\texttt{iii}) The performance improves as regularization strength increases until they fail. Will it even improve if it does not fail?
We study these questions throughout this paper.

\paragraph{\textit{Learning fails when going beyond a \textit{tolerance level} of regularization strength.}}

In order to understand this phenomenon in depth, we show training loss and gradients in Figure \ref{fig:gradients}.\footnote{We observed similar patterns in other networks and data sets in our experiments.}
The training loss excluding regularization penalty in Figure \ref{fig:training_loss} shows that the model learns faster with stronger regularization up to $\lambda=1\times10^{-3}$, but training loss does not decrease at all when even stronger regularization ($\lambda=2\times10^{-3}$) is imposed.
This means there exists a \textit{tolerance level} of regularization strength, which decides success or failure of entire learning.
Gradients on parameters from $\mathcal{L}$ are shown in \ref{fig:avg_abs_grad} to see how such loss does not result in learning.
Compared to less strong regularization ($\lambda=1\times10^{-3}$), the average $|\nabla \mathcal{L}|$ by stronger regularization ($\lambda=2\times10^{-3}$) is much smaller and seems stagnant.
A close-up view with the gradients in logarithmic scale for the first 20 epochs is depicted in Figure \ref{fig:avg_abs_grad_log}.
In a couple of epochs, the models with less strong L1 and L2 regularization start to obtain gradients that are two orders of magnitude greater than their initial gradients.
On the other hand, models with stronger regularization fail to obtain such large gradients, and the magnitudes of gradients rather decay \textit{exponentially}.
Finally, a direct comparison between $|\nabla \mathcal{L}|$ and $\lambda |\nabla \Omega|$ is shown if Figure \ref{fig:grad_portion}, where the total gradient becomes almost completely dominated by $\lambda |\nabla \Omega|$.
The failure in learning can be explained by the following mechanism:
\begin{itemize}
    \item If the regularization is too strong, $\lambda |\nabla \Omega|$ also becomes large.
    \begin{itemize}
        \item If $|\nabla \mathcal{L}| < \lambda |\nabla \Omega|$, we have smaller $|w|$ through the weight update in equation \eqref{eq:gd}. For small $|w|$,  $|\nabla \mathcal{L}|$ is far suppressed while $\lambda |\nabla \Omega|$ is suppressed linearly with $w$ (L2) or constantly (L1). The iterative weight updates can make $\lambda |\nabla \Omega|$ dominate over $|\nabla \mathcal{L}|$, yielding failure in learning.
        \item Otherwise, $|w|$ may not decrease.
    \end{itemize}
\end{itemize}
We explain why small $|w|$ by strong regularization implies far suppressed $|\nabla \mathcal{L}|$ in the next paragraph.

\paragraph{\textit{Why does $|\nabla \mathcal{L}|$ decrease so fast with strong regularization?}}

It is not difficult to see why $|\nabla \mathcal{L}|$ decreases so fast when the regularization is strong. In deep neural networks, the gradients are dictated by back-propagation. It is well known that 
the gradients at the $l^{\textrm{th}}$ layer are given by
\begin{eqnarray}
\frac{\partial \mathcal{L}}{\partial\, \mathbf{w}^{(l)}} = \boldsymbol{\delta}^{(l)}\, (\,\mathbf{a}^{(l-1)}\,)^\intercal
\end{eqnarray}
where $\mathbf{a}^{(l-1)}$ is the output of the neurons at the $(l-1)^{\textrm{th}}$ layer and 
$\boldsymbol{\delta}^{(l)}$ is the $l^{\textrm{th}}$-layer residual which follows the recursive relation
\begin{eqnarray}
\boldsymbol{\delta}^{(l)} = (\,\mathbf{w}^{(l+1)}\,)^\intercal \, \boldsymbol{\delta}^{(l+1)} \,\odot \, \mathbf{a'}^{(l)} 
\end{eqnarray}
where $\odot$ and $\mathbf{a'}$ denote the element-wise multiplications and derivatives of the activation function respectively.

Using the recursive relation, we obtain
\begin{equation}\label{gradient}
\scriptsize
\begin{aligned}
\frac{\partial \mathcal{L}}{\partial \mathbf{w}^{(l)}} &= 
(\,\mathbf{w}^{(l+1)}\,)^\intercal\,
(\,\mathbf{w}^{(l+2)}\,)^\intercal\, \cdots \,
(\,\mathbf{w}^{(L)}\,)^\intercal \, \boldsymbol{\delta}^{(L)}  \\
& \odot\, 
\mathbf{a'}^{(L-1)} \, \odot\,
\mathbf{a'}^{(L-2)} \, \odot\,
\cdots \, \odot\,
\mathbf{a'}^{(l+1)} \, \odot\,
\mathbf{a'}^{(l)} \, (\,\mathbf{a}^{(l-1)}\,)^\intercal
\end{aligned}
\normalsize
\end{equation}

If the regularization is too strong, the weights would be significantly suppressed with penalty in \eqref{eq:gd}, which is also observed in Figure \ref{fig:weight_comp}. From \eqref{gradient}, since the gradients are proportional to the product of the weights at later layers (whose magnitudes are typically much less than 1, particularly in the beginning of training \cite{glorot2010understanding}), they are even more suppressed. 

In fact, the suppression is more severe than what we have deduced above. The factor $\mathbf{a}^{(l-1)}$ in \eqref{gradient} could actually lead to further suppression to the gradients when the weights are very small, for the following reasons. First of all, we use ReLU as the activation function and it could be written as
\begin{eqnarray} \label{eq:relu}
\textrm{ReLU}(x) = x\,\, \Theta(x)  
\end{eqnarray}
where $\Theta(x)$ is the Heaviside step function. Using this, we could write
\begin{eqnarray}\label{eq:relu2}
\label{relu}
\mathbf{a}^{(l-1)} = \left(\,\mathbf{w}^{(l-1)}\,\mathbf{a}^{(l-2)}\,\right)\,
\odot\, \Theta\left(\, \mathbf{w}^{(l-1)}\,\mathbf{a}^{(l-2)} \,\right)
\end{eqnarray}

Applying \eqref{relu} recursively, we can see that $\mathbf{a}^{(l-1)}$ is proportional to the product of the weights at previous layers. Again, when the weights are suppressed by strong regularization, $\mathbf{a}^{(l-1)}$ would be suppressed correspondingly. Putting everything together, we can conclude that in the presence of strong regularization, 
the gradients are far more suppressed than the weights.

Strictly speaking, the derivations above are valid only for fully-connected layers. For convolutional layers, the derivations are more complicated but similar. Our conclusions above would still be valid.

\paragraph{\textit{Discussion}} 
Please note that this phenomenon is different from vanishing gradients caused by weight initialization or saturating activation functions such as sigmoid and hyperbolic tangent, which typically make learning slow and are significantly relieved by ReLU, a non-saturating activation function.
In contrast, strong regularization does not even let learning start, and the symptom worsens as training proceeds, which makes learning completely fail.
In addition, ReLU is adopted for both baselines and our approaches in the experiments.

In order to claim that the sudden failure also follows from vanishing gradients in deep networks, we would at least need to know that the gradients would ``suddenly'' vanish as the regularization strength increases.
To the best of our knowledge, there is no such an analysis, be it theoretical or empirical.
In addition, using equation \eqref{eq:relu} and applying equation \eqref{eq:relu2} recursively, we show that vanishing weights lead to another level of suppression on the gradients, which is novel.

\subsection{Gradient-Coherent Strong Regularization}\label{sec:approach} 

In order to prevent failure in learning, we propose \textit{gradient-coherent strong regularization} to selectively impose strong regularization only when the gradients from regularization ($\frac{\partial \Omega}{\partial \mathbf{w}}$) do not obstruct learning too much.
By comparing $\frac{\partial \mathcal{L}}{\partial \mathbf{w}}$ with $ \frac{\partial \mathcal{L}}{\partial \mathbf{w}}+\lambda \frac{\partial \Omega}{\partial \mathbf{w}}$, we can find out how the gradients from regularization affect the overall learning, according to equation \eqref{eq:gd}.
That is, we measure quality of regularization gradients for $\frac{\partial \mathcal{L}}{\partial \mathbf{w}}$, where we assume that there will be no learning if their quality is sufficiently low.
Thus, we define the regularization strength at step $t$ as
\begin{equation}\label{eq:gg2}
	\begin{aligned}
	\lambda^{(t)} &=
    \begin{cases}
        \lambda & \text{if } \pi^{(t)} > \mu \\
        0 &\text{otherwise}\\
    \end{cases}
    \end{aligned}
\end{equation}
where $\pi^{(t)}$ is the quality of $\frac{\partial \Omega}{\partial \mathbf{w}}$ for $\frac{\partial \mathcal{L}}{\partial \mathbf{w}}$, and $\mu$ is a hyper-parameter.
Only when the quality is high enough, the regularization is imposed.

Meanwhile, it is rather difficult to measure the actual interference by regularization  for each weight.
Even when there is a great magnitude change in a gradient after adding the gradient from strong regularization, if the resulting gradient keeps the same sign, the change may still be useful for learning and may even accelerate learning.
On the other hand, it is obvious that the resulting gradient with opposite sign is harmful.
Hence, we propose a gradient sign coherence rate to approximate coherence between $\frac{\partial \mathcal{L}}{\partial \mathbf{w}}$ and $ \frac{\partial \mathcal{L}}{\partial \mathbf{w}}+\lambda \frac{\partial \Omega}{\partial \mathbf{w}}$, to measure quality of $\frac{\partial \Omega}{\partial \mathbf{w}}$.
It is defined as
\begin{equation}\label{eq:gsar}
	\begin{aligned}
	\pi^{(t)} = \left. \frac{
	|| \Theta (\text{sign}( \frac{\partial \mathcal{L}}{\partial \mathbf{w}} ) \, \odot \,
	\text{sign}( \frac{\partial \mathcal{L}}{\partial \mathbf{w}} + 
	    \lambda \frac{\partial \Omega}{\partial \mathbf{w}} ) ) ||_1
	}{n}  \right|_{\mathbf{w}=\mathbf{w}^{(t)}}
    \end{aligned}
\end{equation}
where $n$ is the number of parameters, and $\pi^{(t)}$ lies in [0,1], where $\pi^{(t)}=1$ means a complete coherence.
Thus, a sign coherence rate between two random vectors is expected to be 0.5.\footnote{In practice, when we compute $\pi$, the parameters with $\nabla \mathcal{L}=0$ are excluded because we want to measure coherence when there is learning, \textit{i.e.}, $\nabla \mathcal{L} \neq 0$.}
What the proposed approach does is that it measures how much the enforcement of regularization will change the direction of $\frac{\partial \mathcal{L}}{\partial \mathbf{w}}$ and it enforces regularization only if the direction is not too much changed.
The computation of $\pi$ requires only a couple of vector computations, which can be done efficiently with GPUs.

\begin{figure}[tbh] 
\centering
\begin{subfigure}{.5\columnwidth}
    \centering
    \includegraphics[width=0.9\linewidth]{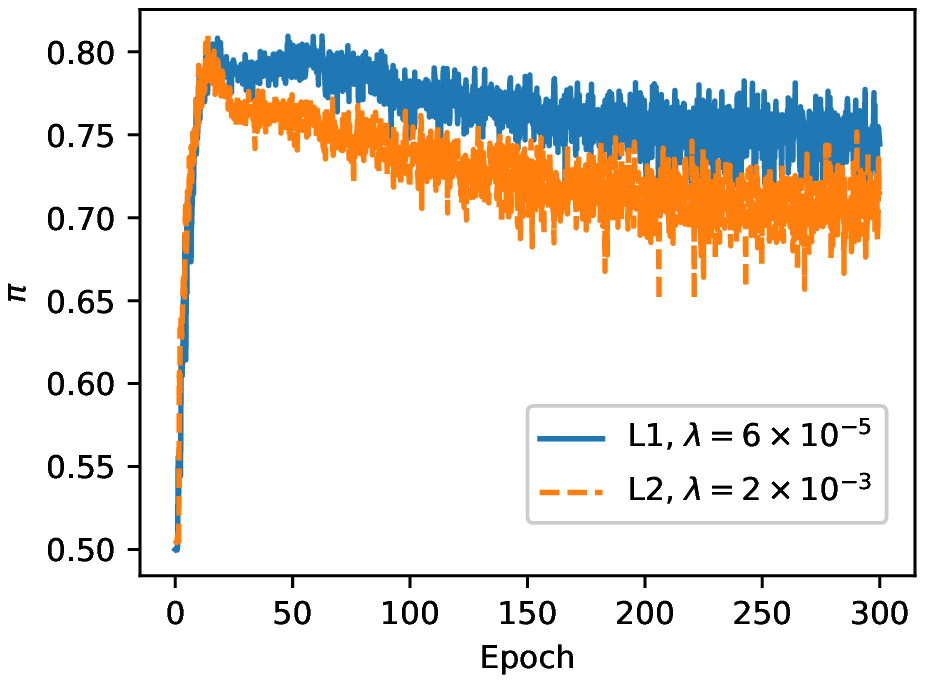}
\end{subfigure}%
\begin{subfigure}{.5\columnwidth}
    \centering
    \includegraphics[width=0.9\linewidth]{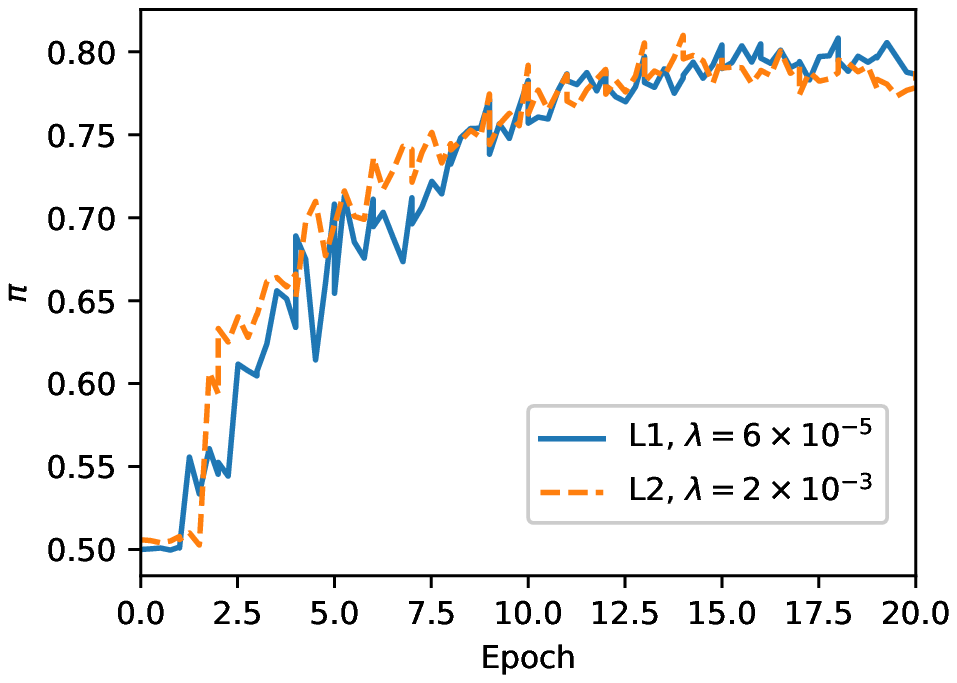}
\end{subfigure}%
\caption{ Gradient sign coherence rate (left) and its close-up view (right), by VGG-16 on CIFAR-100. }
\label{fig:gsar}
\end{figure}

The example of the gradient sign coherence rate for strong regularization is depicted in Figure \ref{fig:gsar}, where $\lambda \frac{\partial \Omega}{\partial \mathbf{w}}$ is used only to compute the rate and is not added for learning.
Indeed, $\pi$ is about 0.5 in the first couple of epochs, which means that the gradients would be greatly affected by strong regularization, and it quickly increases for the next 20 epochs.
Then, with the convergence of the model, it decreases a bit, which is reasonable.

\paragraph{\textit{Proximal gradient algorithm for L1 regularizer}}
Meanwhile, since L1 norm is not differentiable at zero, we employ the proximal gradient algorithm \cite{parikh2014proximal}, which enables us to obtain proper sparsity (\textit{i.e.,} guaranteed convergence) for non-smooth regularizers.
We use the following update formulae:
\begin{equation}\label{eq:pgd}
    \small
	\begin{aligned}
	w^{(t')} &= w^{(t)} -  \alpha \left.
	\frac{\partial \mathcal{L}}{\partial w} \right|_{w=w^{(t)}}\\
	w^{(t+1)} &= \text{prox}_{\alpha\lambda\Omega} (w^{(t')}) =
	S(w^{(t')}, \alpha \lambda^{(t)})\\
	S(a,z) &=
	\begin{cases}
    a-z &\text{if } a > z\\
    a+z &\text{if } a < -z\\
    0 &\text{otherwise.}
    \end{cases}
    \end{aligned}
    \normalsize
\end{equation}

\paragraph{Discussion}
Normalization techniques such as batch normalization \cite{ioffe2015batch} and weight normalization \cite{salimans2016weight} can be possible approaches to prevent $\nabla \mathcal{L}$ from diminishing quickly.
However, it has been shown that L2 regularization has no regularizing effect when combined with normalization but only influences on the effective learning rate \cite{van2017l2}.
In other words, the normalization techniques do not actually simplify the solution since the decrease in parameter magnitude is canceled by normalization.
This does not meet our goal, which is to heavily simplify solutions to reduce overfitting and compress networks.
It is also worth mentioning that normalization techniques can be taken out from DNNs without losing test accuracy if the model is initialized properly and regularized well \cite{zhang2019fixup}.

\section{Experiments}\label{sec:experiments}
\begin{table}[tbh]
\small
\begin{center}
\begin{tabular}{cccc}
\multirow{2}{*}{Dataset} & \multirow{2}{*}{Classes} & Training Images & Test Images  \\
& &  per Class & per Class\\
\hline
CIFAR-10 & 10 & 5000 & 1000 \\
CIFAR-100 & 100 & 500 & 100 \\
SVHN &  10 & 7325.7 (avg.) & 2603.2 (avg.) \\
\end{tabular}
\end{center}
\caption{Dataset statistics for CIFAR-10 and CIFAR-100.}
\label{tab:data}
\normalsize
\end{table}
We first evaluate the effectiveness of our proposed method with popular architectures, AlexNet and VGG-16 on the public datasets CIFAR-10 and CIFAR-100 \cite{krizhevsky2009learning}.
Then, we employ variations of VGG on another public dataset SVHN \cite{netzer2011reading}, in order to see the effect of the number of hidden layers on the tolerance level for strong regularization.
Please note that we do not employ recent architectures that contain normalization techniques such as batch normalization \cite{ioffe2015batch}, for the reason described in the previous section.\footnote{As it was recently shown that such architectures without normalization techniques can perform very well with proper initialization and regularization \cite{zhang2019fixup}, our strong regularization with the proper initialization is left as our future work.}
All the datasets contain images of 32$\times$32 resolution with 3 color channels.
The dataset statistics are described in Table \ref{tab:data}.
AlexNet and VGG-16 for CIFAR-10 and CIFAR-100 contain 2.6 and 15.2 million parameters, respectively.
VGG-11 and VGG-19 contain 9.8 and 20.6 millions of parameters, respectively.

L1/L2 regularization is applied to all network parameters except bias terms.
We use PyTorch\footnote{\url{http://pytorch.org/}} framework for all experiments, and we use its official computer vision library\footnote{\url{https://github.com/pytorch/vision}} for the implementations of the networks.
In order to accommodate the datasets, we made some modifications to the networks.
The kernel size of AlexNet's max-pooling layers is modified from 3 to 2, and the first convolution layer's padding size is modified from 2 to 5.
All of its fully connected layers are modified to have 256 neurons.
For VGG, we modified the fully connected layers to have 512 neurons.
The output layers of both networks have 10 neurons for CIFAR-10 and SVHN, and 100 neurons for CIFAR-100.
The networks are learned by stochastic gradient descent with momentum of 0.9.
The batch size is set to 128, and the initial learning rate is set to 0.05 and decayed by a factor of 2 every 30 epochs.
We use dropout layers (with drop probability 0.5) and pre-process training data\footnote{We apply horizontal random flipping and random cropping to original images in each batch. We do not apply them to SVHN as they may harm the performance.} in order to report the extra performance boost on top of common regularization techniques.

AlexNet and VGG-16 are experimented for different regularization methods (L1 and L2) and different datasets (CIFAR-10 and CIFAR-100), yielding 8 experiment sets.
Then, VGG-11, VGG-16, and VGG-19 are experimented for L1 and L2 regularization methods on SVHN, yielding 6 experiment sets.
For each experiment set, we set the \textbf{baseline} method as the one with best-tuned L1 or L2 regularization but without our gradient-coherent regularization.
For each of L1 and L2 regularization, we try more than 10 different values of $\lambda$, and for each $\lambda$, we report average accuracy of three independent runs and report 95\% confidence interval.
We perform statistical significance test (t-test) for the improvement over the baseline method.
We also report \textbf{sparsity} of each trained model, which is the proportion of the number of zero-valued parameters to the number of all parameters.
Please note that we mean the sparsity by the one actually derived by the models, not by pruning parameters with threshold after training.

In this specific task, for a clear comparison with the baseline, a fixed amount of regularization needs to be imposed without skipping arbitrary many training steps, to reach a similar level of smoothness or sparsity in the solution.
We observe in Figure \ref{fig:avg_abs_grad_log} and \ref{fig:gsar} that the average $|\nabla \mathcal{L}|$ is elevated two orders of magnitude and the gradient sign coherent rate quickly drifts from 0.5 (a random coherence) in the first couple of epochs.
Therefore, we also employ the following regularization schedule:
\begin{equation}\label{eq:gg}
    \begin{aligned}
    \lambda^{(t)} &=
    \begin{cases}
        \lambda &\text{if epoch}(t) \geq \gamma\\
        0 &\text{otherwise}
    \end{cases}
    \end{aligned}
\end{equation}
where epoch$(t)$ is the epoch number of the time step $t$, and $\gamma$ is a hyper-parameter that is determined by $\pi$.
The formula means that we do not impose any regularization until $\gamma^{\textrm{th}}$ epoch, and then impose strong regularization until training ends.
We call this approach \textbf{ours} and employ it for the majority of the experiments, but we also experiment with the original approach in equation \eqref{eq:gg2}, which we call \textbf{ours\_orig}.
Please note that \textbf{ours\_orig} is superior to \textbf{ours} because it can avoid over-regularization at any training steps.
Considering dynamics of stochastic gradient descent, we can set the starting point $\gamma$ as the time step where $\pi$ becomes a little greater than 0.5.
We set $\gamma=5$, where $\pi$ reaches 0.7 in general.
We did not find significantly different results for $2 \leq \gamma \leq 20$.

\subsection{Results on CIFAR-10 and CIFAR-100}

\begin{figure*}[htb] 
\centering
\begin{subfigure}{.25\textwidth}
    \centering
    \includegraphics[width=0.9\linewidth]{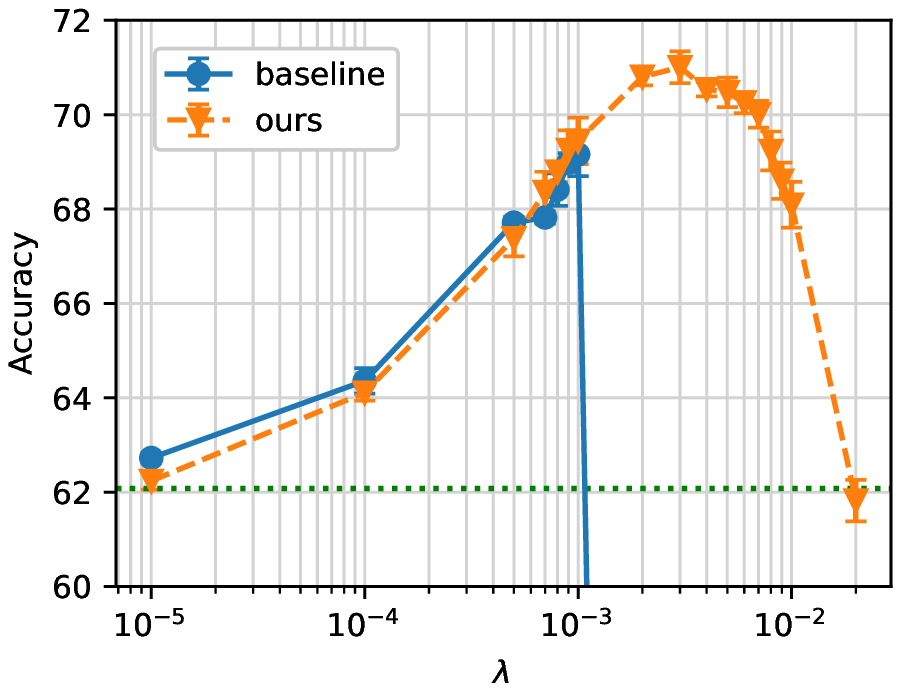}
    \caption{VGG-16, CIFAR-100, L2}
    \label{fig:vgg_res1}
\end{subfigure}%
\begin{subfigure}{.25\textwidth}
    \centering
    \includegraphics[width=0.9\linewidth]{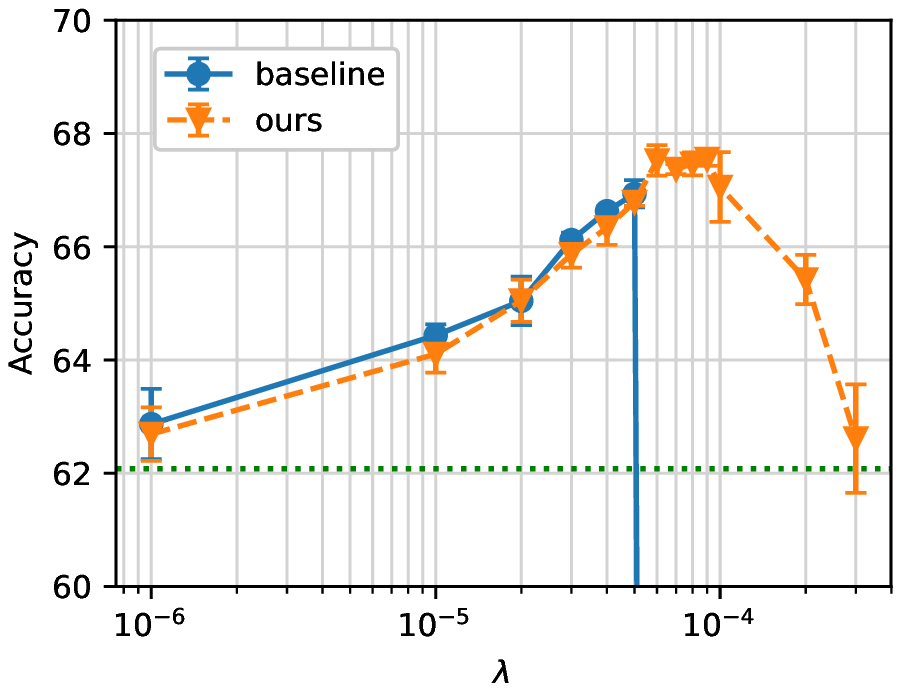}
    \caption{VGG-16, CIFAR-100, L1}
    \label{fig:vgg_res2}
\end{subfigure}%
\begin{subfigure}{.25\textwidth}
    \centering
    \includegraphics[width=0.9\linewidth]{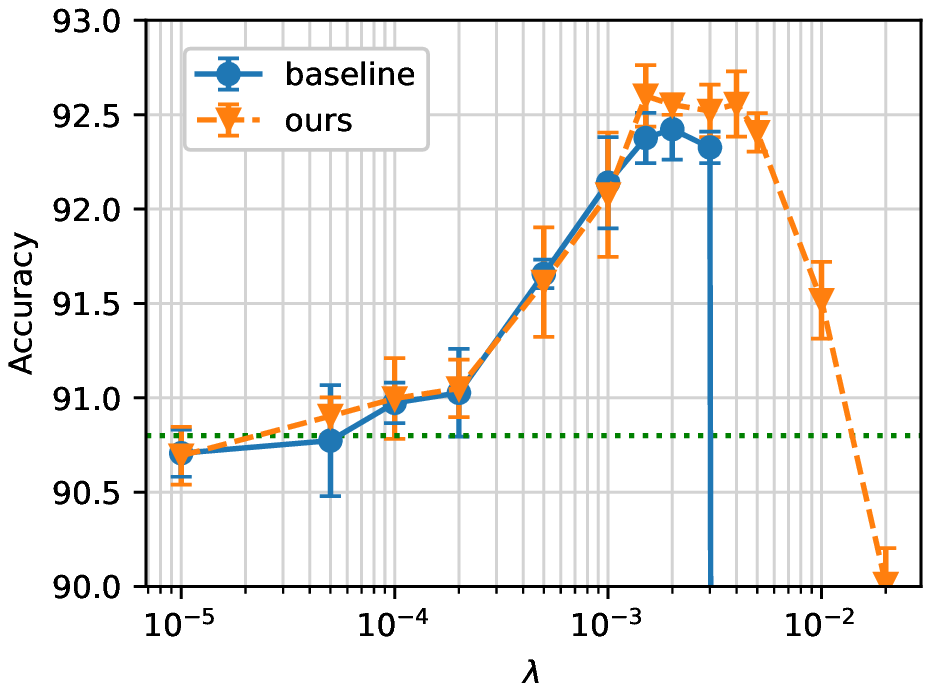}
    \caption{VGG-16, CIFAR-10, L2}
    \label{fig:vgg_res4}
\end{subfigure}%
\begin{subfigure}{.25\textwidth}
    \centering
    \includegraphics[width=0.9\linewidth]{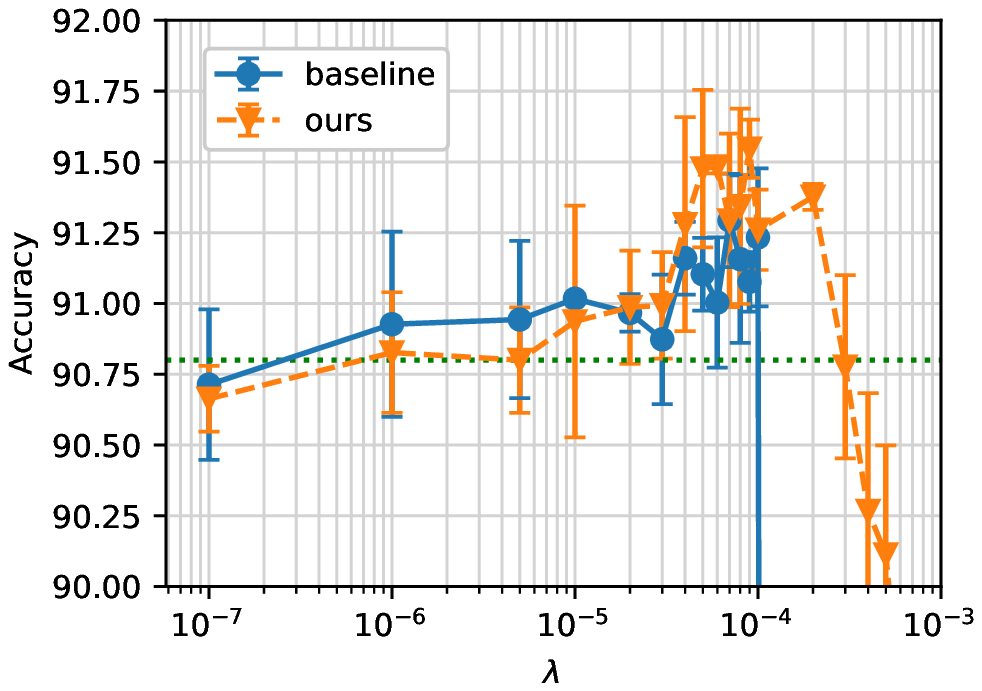}
    \caption{VGG-16, CIFAR-10, L1}
    \label{fig:vgg_res5}
\end{subfigure}
\begin{subfigure}{.25\textwidth}
    \centering
    \includegraphics[width=0.9\linewidth]{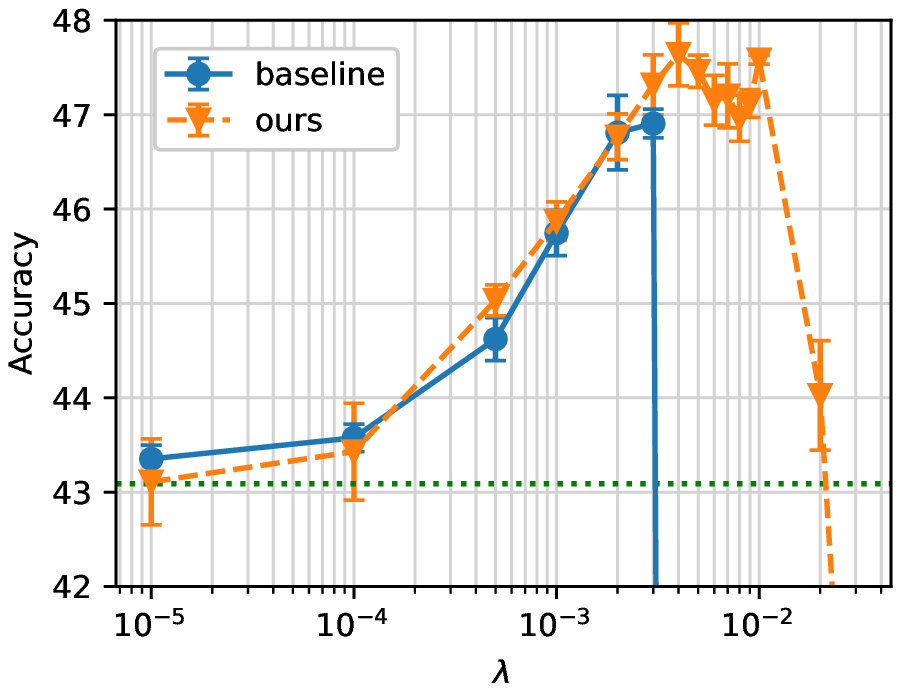}
    \caption{AlexNet, CIFAR-100, L2}
    \label{fig:alex_res1}
\end{subfigure}%
\begin{subfigure}{.25\textwidth}
    \centering
    \includegraphics[width=0.9\linewidth]{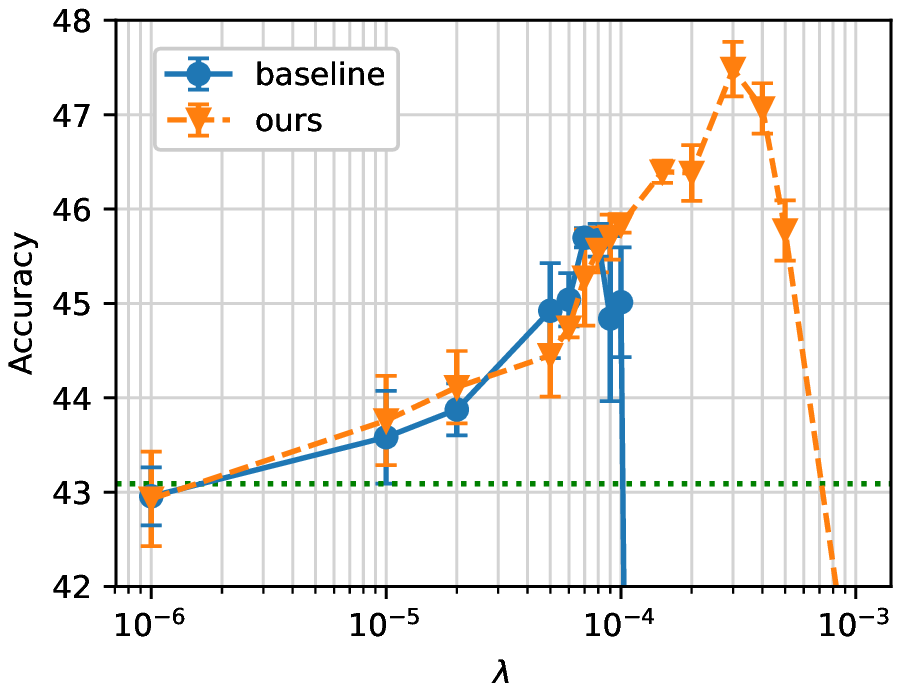}
    \caption{AlexNet, CIFAR-100, L1}
    \label{fig:alex_res2}
\end{subfigure}%
\begin{subfigure}{.25\textwidth}
    \centering
    \includegraphics[width=0.9\linewidth]{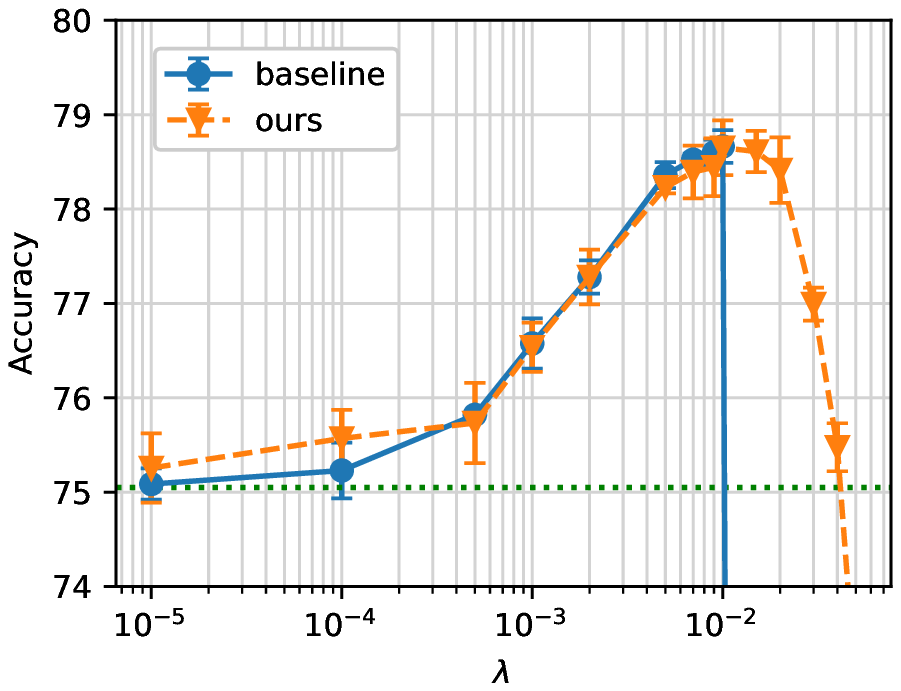}
    \caption{AlexNet, CIFAR-10, L2}
    \label{fig:alex_res4}
\end{subfigure}%
\begin{subfigure}{.25\textwidth}
    \centering
    \includegraphics[width=0.9\linewidth]{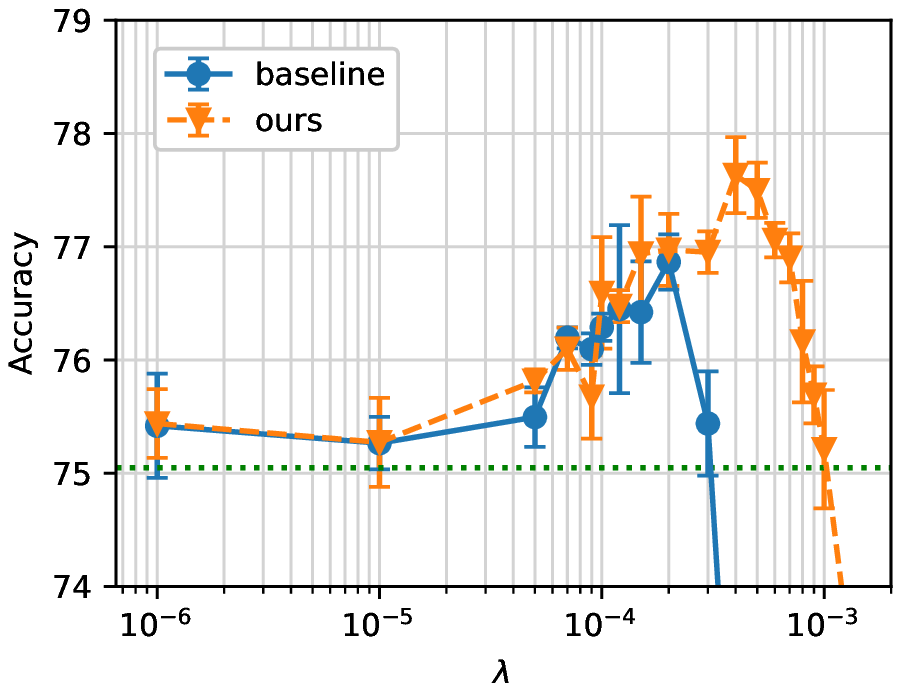}
    \caption{AlexNet, CIFAR-10, L1}
    \label{fig:alex_res5}
\end{subfigure}
\caption{ Accuracy obtained by VGG-16 (a,b,c,d) and AlexNet (e,f,g,h). A green dotted horizontal line is accuracy obtained by a model without L1/L2 regularization (but with dropout). The error bars indicate 95\% confidence interval.}
\label{fig:vgg_res}
\end{figure*}

\begin{figure*}[htb] 
\centering
\begin{subfigure}{.25\textwidth}
    \centering
    \includegraphics[width=0.9\linewidth]{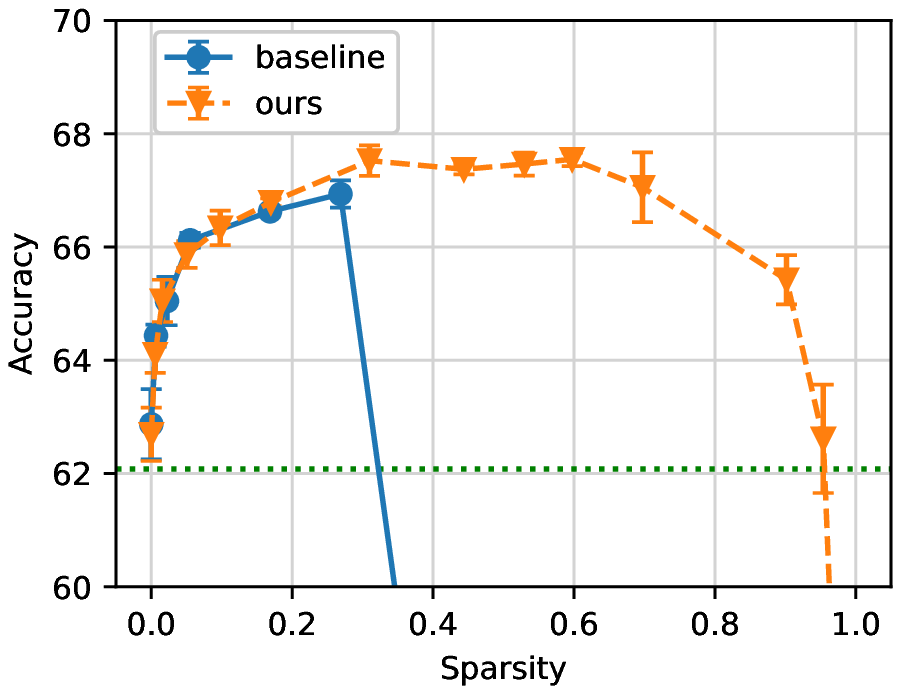}
    \caption{VGG-16, CIFAR-100, L1}
    \label{fig:vgg_res3}
\end{subfigure}%
\begin{subfigure}{.25\textwidth}
    \centering
    \includegraphics[width=0.9\linewidth]{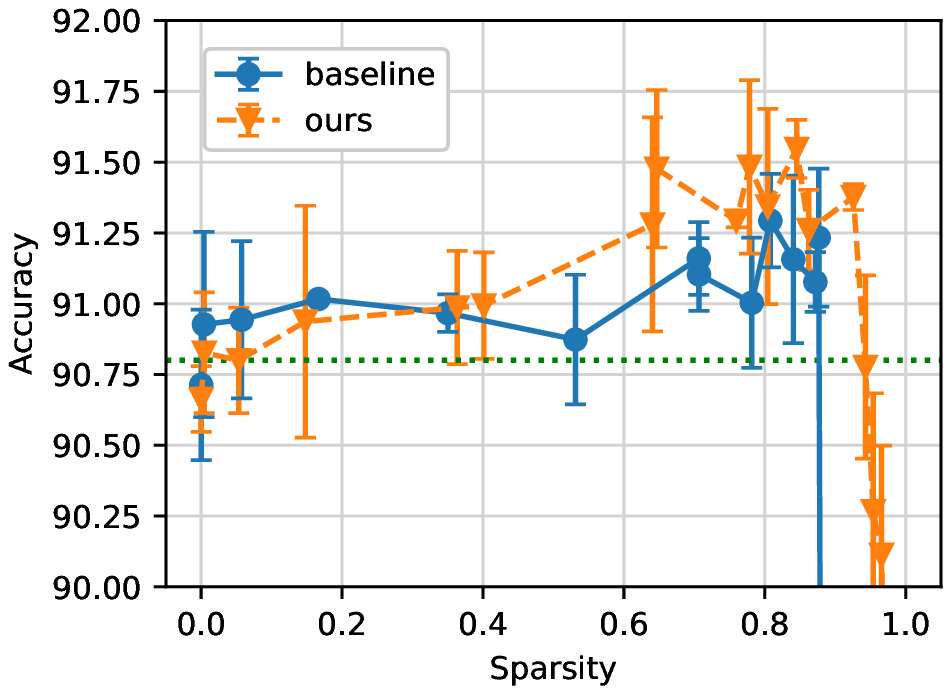}
    \caption{VGG-16, CIFAR-10, L1}
    \label{fig:vgg_res6}
\end{subfigure}%
\begin{subfigure}{.25\textwidth}
    \centering
    \includegraphics[width=0.9\linewidth]{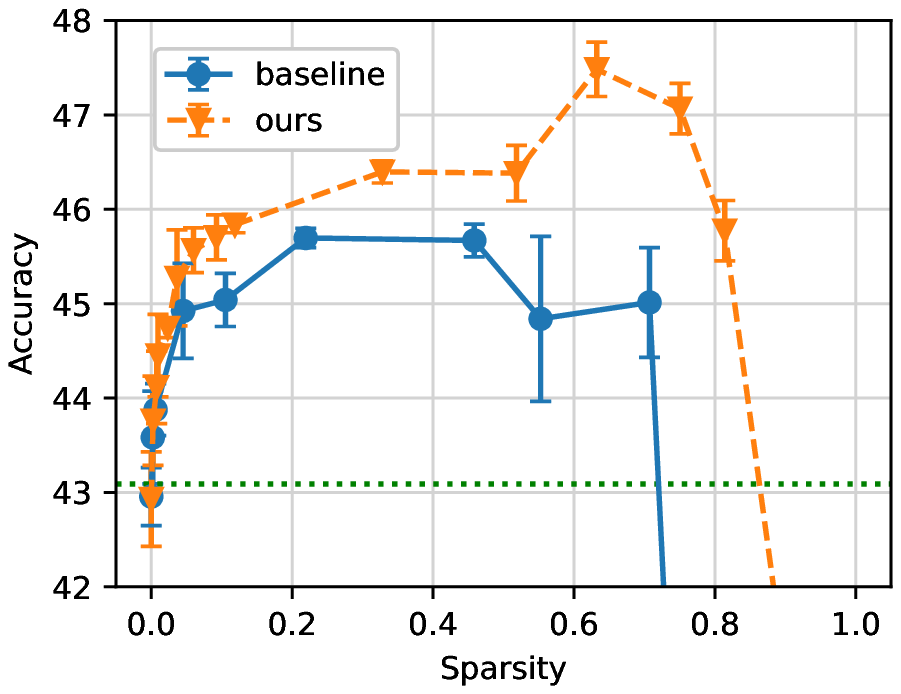}
    \caption{AlexNet, CIFAR-100, L1}
    \label{fig:alex_res3}
\end{subfigure}%
\begin{subfigure}{.25\textwidth}
    \centering
    \includegraphics[width=0.9\linewidth]{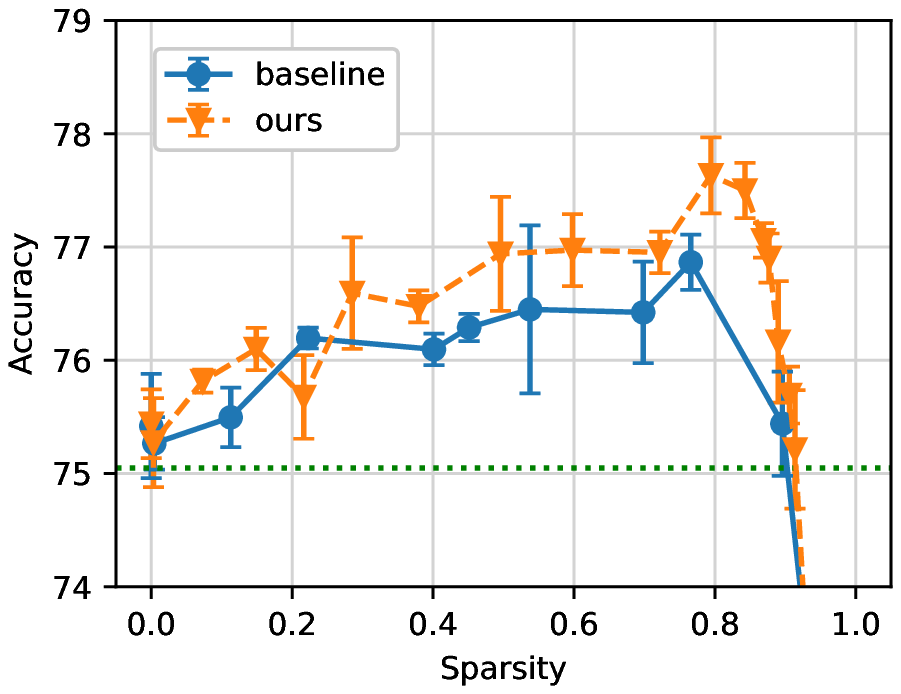}
    \caption{AlexNet, CIFAR-10, L1}
    \label{fig:alex_res6}
\end{subfigure}
\caption{ Accuracy for different sparsity when L1 regularization is employed. The error bars indicate 95\% confidence interval.}
\label{fig:alex_res}
\end{figure*}

The experimental results on CIFAR-10 and CIFAR-100 are depicted in Figure \ref{fig:vgg_res} and \ref{fig:alex_res}.
As we investigated in the previous section, the baseline method suddenly fails beyond certain values of tolerance level.
However, our proposed method does not fail for higher values of $\lambda$, and it indeed achieves higher accuracy in general.
Another interesting observation is that, unlike VGG-16, we obtain more improvement by AlexNet  with L1 regularization.
Meanwhile, tuning with the regularization parameter can be difficult as the curves have somewhat sharp peak, but our proposed method ease the problem to some extent by preventing the sharp drop, \textit{i.e.}, the sudden failure.
Our L1 regularizer obtains better sparsity for the similar level of accuracy (Figure \ref{fig:alex_res}), which means that strong regularization is promising for compression of DNNs.
Overall, the improvement is more prominent on CIFAR-100 than on CIFAR-10, and we think this is because overfitting can more likely occur on CIFAR-100 as there are fewer images per class in CIFAR-100 than in CIFAR-10.

Interestingly, our proposed method often obtains higher accuracy even when the baseline does not fail on CIFAR-10, and this is prominent especially when $\lambda$ is a little less than the tolerance level (better shown in Figure \ref{fig:vgg_res6}, \ref{fig:alex_res3}, \ref{fig:alex_res6}).
One possible explanation is that avoiding strong regularization in the early stage of training can help the model to explore the parameter space more freely, and the better exploration results in finding superior local optima.

\begin{table}[htb]
\small
\begin{center}
\begin{tabular}{lcc}
& CIFAR-100 & CIFAR-10 \\
\hline
\textbf{VGG-16} & & \\
No L1/L2 & 62.08\scriptsize{$\pm$0.81} & 90.80\scriptsize{$\pm$0.23} \\
\cdashline{1-3}
L2 baseline & 69.16\scriptsize{$\pm$0.46} & 92.42\scriptsize{$\pm$0.16}\\
L2 ours & 71.01\scriptsize{$\pm$0.33} & 92.60\scriptsize{$\pm$0.16}\\
Rel. improvement & \textbf{+2.67\%} & +0.19\% \\
\cdashline{1-3}
L1 baseline & 66.94\scriptsize{$\pm$0.24} & 91.29\scriptsize{$\pm$0.16}\\
L1 ours & 67.55\scriptsize{$\pm$0.12} & 91.55\scriptsize{$\pm$0.10}\\
Rel. improvement & \textbf{+0.91\%}  & \textbf{+0.28\%} \\
\hline
\textbf{AlexNet} & & \\
No L1/L2 & 43.09\scriptsize{$\pm$0.25} & 75.05\scriptsize{$\pm$0.20}\\
\cdashline{1-3}
L2 baseline & 46.91\scriptsize{$\pm$0.15} & 78.66\scriptsize{$\pm$0.17}\\
L2 ours & 47.64\scriptsize{$\pm$0.33} & 78.65\scriptsize{$\pm$0.29}\\
Rel. improvement & \textbf{+1.56\%} & -0.01\% \\
\cdashline{1-3}
L1 baseline & 45.70\scriptsize{$\pm$0.10} & 76.87\scriptsize{$\pm$0.24}\\
L1 ours & 47.48\scriptsize{$\pm$0.29} & 77.63\scriptsize{$\pm$0.34}\\
Rel. improvement & \textbf{+3.89\%} & \textbf{+0.99\%} \\
\hline
\end{tabular}
\end{center}
\caption{Accuracy with 95\% confidence interval. Note that when no L1/L2 regularization is imposed, dropout is still employed. Statistically significant improvements are bold-faced.}
\label{tab:overall}
\normalsize
\end{table}

The exact accuracy obtained is shown in Table \ref{tab:overall}.
Our proposed model always improves the baselines by up to 3.89\%, except AlexNet with L1 regularization on CIFAR-10, and most (6 out of 7) improvements are statistically significant.
Also, L1/L2 regularization seems indeed useful even when dropout is employed; our model improves the baseline that is without L1 or L2 regularization but with dropout, by 14.4\% in accuracy.

\begin{figure}[htb] 
\centering
\begin{subfigure}{.5\columnwidth}
    \centering
    \includegraphics[width=0.9\linewidth]{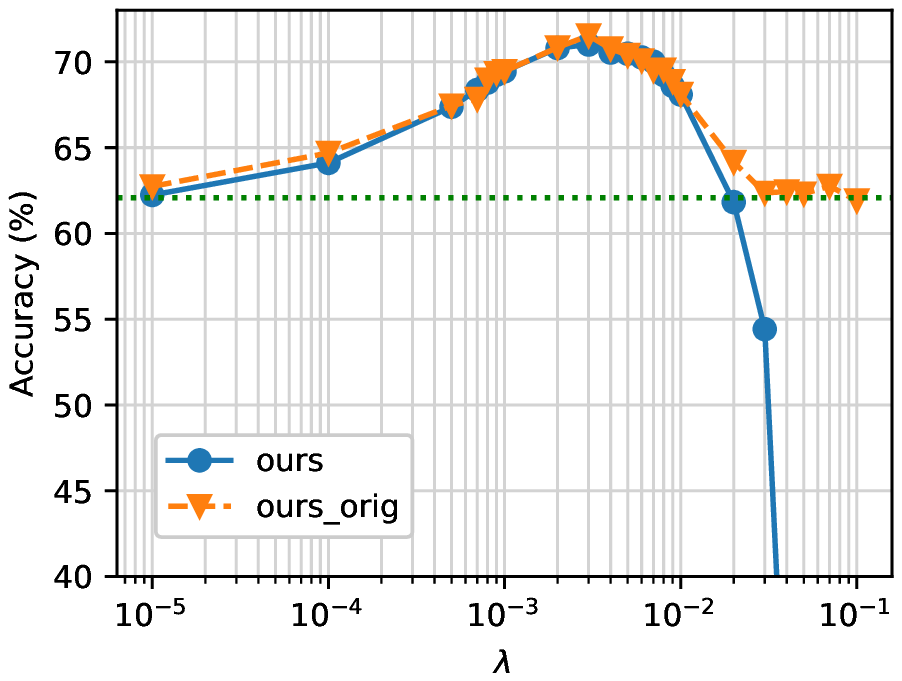}
\end{subfigure}%
\begin{subfigure}{.5\columnwidth}
    \centering
    \includegraphics[width=0.9\linewidth]{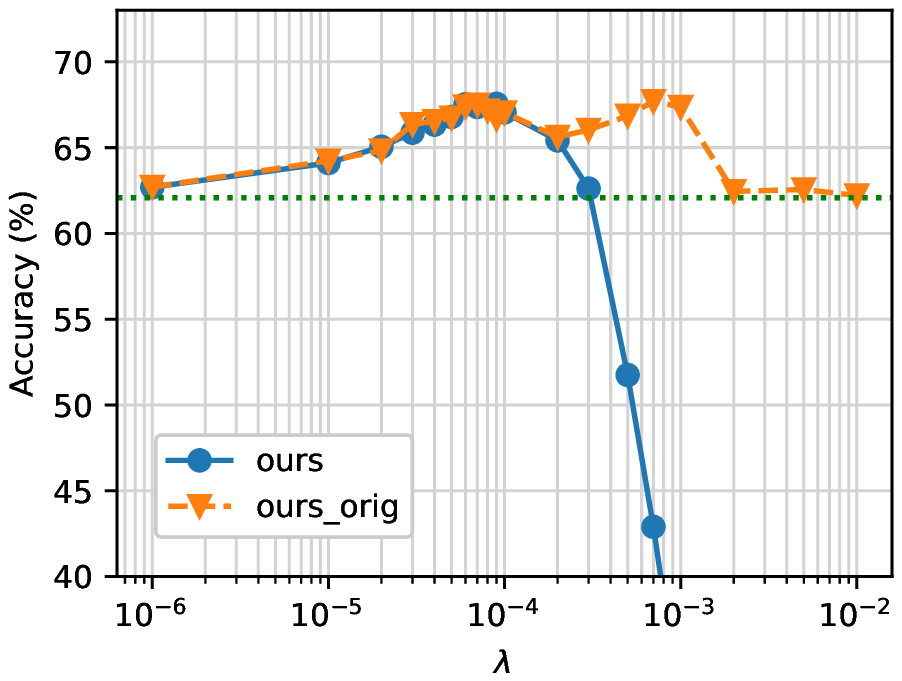}
\end{subfigure}
\caption{Acccuracy on CIFAR-100 by VGG-16 with \textbf{ours} and \textbf{ours\_orig}. L2 regularizer (left) and L1 regularizer (right) are employed. $\mu$ is set to 0.6 for both.}
\label{fig:ours_orig}
\end{figure}

\paragraph{Results by ours\_orig}
We also perform experiments by \textbf{ours\_orig} in equation \eqref{eq:gg2} and compare the results with \textbf{ours}; the results are shown in Figure \ref{fig:ours_orig}. 
Although \textbf{ours} does not suffer from sudden failure in learning by strong regularization, it performs poorly for very strong regularization.
This is because the gradients from regularization are too big so that the overall gradients are too much corrupted.
However, \textbf{ours\_orig} skips regularization if the quality of gradients from regularization is not good enough, so it can still perform well for very strong regularization.
As a result, it can be easier to set $\lambda$ with \textbf{ours\_orig}.
The results are similar for other data sets and architectures and thus are omitted.

\subsection{SVHN Results: Does the Number of Layers Affect the Failure by Strong Regularization? }\label{sec:svhn} 

\begin{figure*}[tbh] 
\centering
\begin{subfigure}{.33\textwidth}
    \centering
    \includegraphics[width=0.8\linewidth]{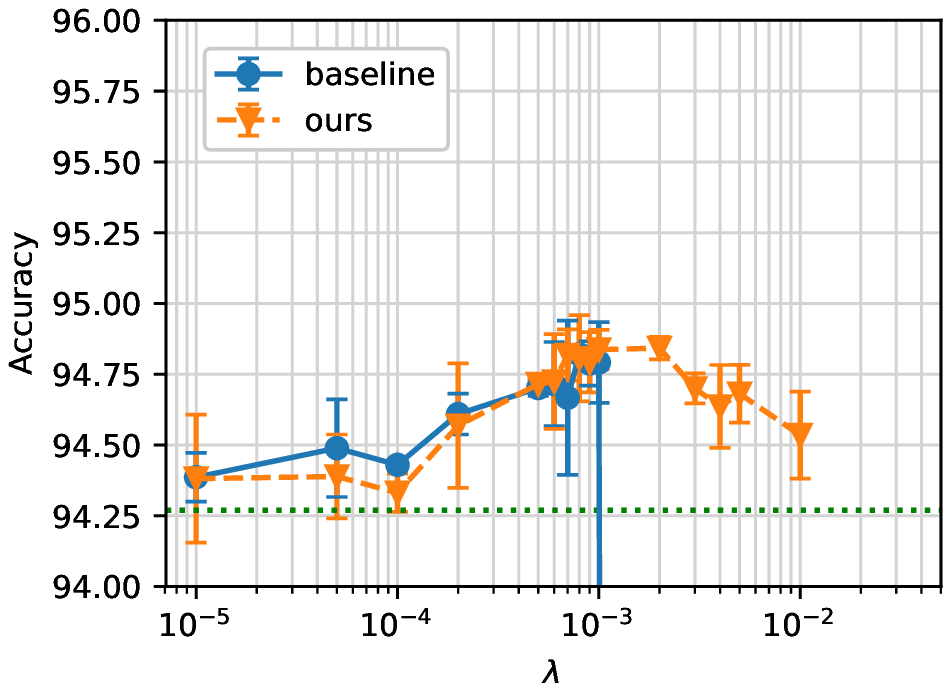}
    \caption{VGG-11}
\end{subfigure}%
\begin{subfigure}{.33\textwidth}
    \centering
    \includegraphics[width=0.8\linewidth]{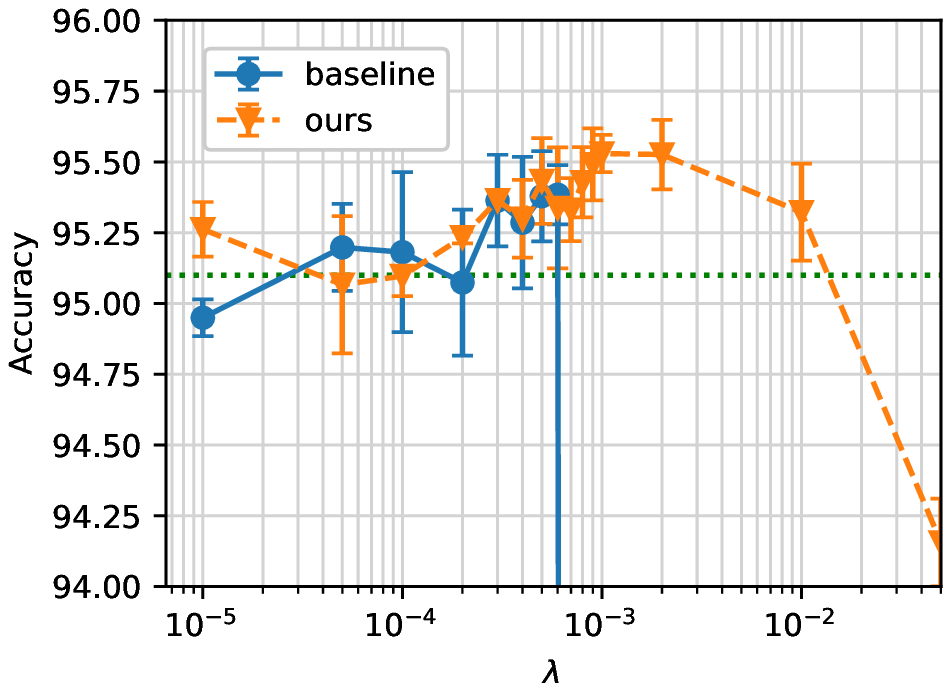}
    \caption{VGG-16}
\end{subfigure}%
\begin{subfigure}{.33\textwidth}
    \centering
    \includegraphics[width=0.8\linewidth]{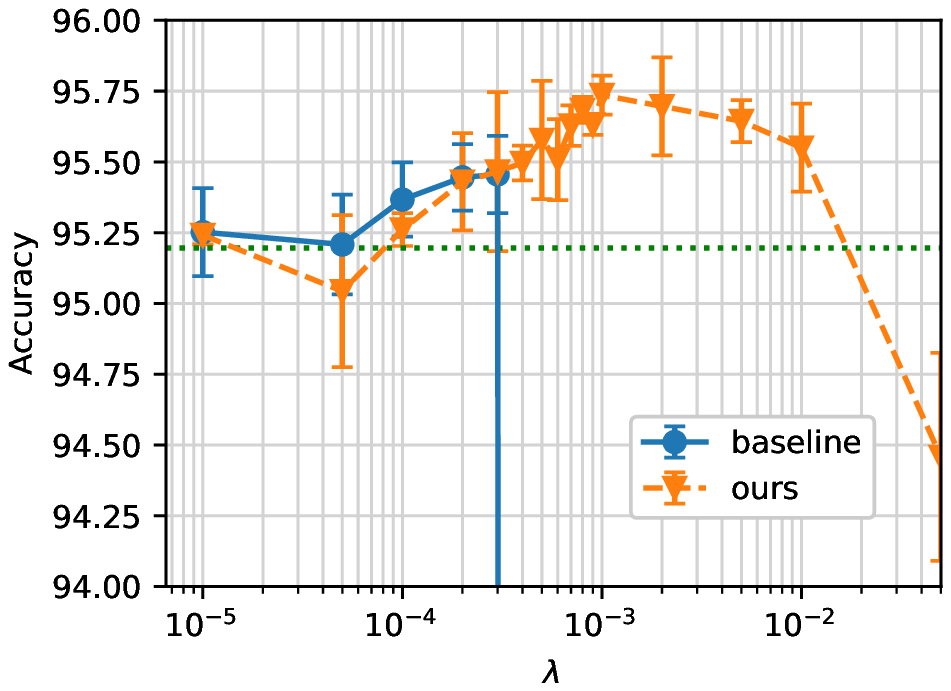}
    \caption{VGG-19}
\end{subfigure}
\caption{ Accuracy obtained by variations of VGG with L2 regularization on SVHN. A green dotted horizontal line is an accuracy obtained by a model without L2 regularization (but with dropout). The error bars indicate 95\% confidence interval.}
\label{fig:svhn}
\end{figure*}

The analysis in Section \ref{sec:fail} implies that the number of hidden layers would affect the tolerance level when strong regularization is imposed.
That is, if there are more hidden layers in the neural network architecture, the learning will  more easily fail by strong regularization.
In order to substantiate the hypothesis empirically, we employ variations of the VGG architecture, \textit{i.e.,} VGG-11, VGG-16, and VGG-19, which consist of 11, 16, and 19 hidden layers, respectively.
Experiments are performed on the SVHN dataset.

The results by L2 regularization are depicted in Figure \ref{fig:svhn}.\footnote{The results by L1 regularization show similar patterns and are omitted due to space limit.}
For all VGG variations, the peaks of our method's accuracy is formed around $\lambda=1\times 10^{-3}$.
As more hidden layers are added to the network, the tolerance level where the baseline suddenly fails is more and more decreased.
This means that deeper architectures are indeed more likely to fail by strong regularization, as hypothesized by our analysis.

Because the method without L1/L2 regularization already performs well on this dataset and there are relatively many training images per class, the improvements by L1/L2 regularization are not big.
Our method still outperforms the baseline in all experiments (6 out of 6), but the improvement is less statistically significant (2 of 6) compared to CIFAR-10 and CIFAR-100 experiments.

\subsection{Network Compression by Strong Regularization} 

\begin{table}[tbh]
\scriptsize
\begin{center}
\begin{tabular}{lccc}
& \multirow{2}{*}{Sparsity} & \multirow{2}{*}{Accuracy} & Compression\\
& & & rate\\
\hline
\multicolumn{4}{l}{\textbf{AlexNet on CIFAR-100}} \\
L1 baseline & 0.219 & 45.70\scriptsize{$\pm$0.10} & \multirow{2}{*}{4.2$\times$} \\
L1 ours (sparse) & 0.814 & 45.77\scriptsize{$\pm$0.32} & \\
\cdashline{1-4}
\multicolumn{4}{l}{\textbf{AlexNet on CIFAR-10}} \\
L1 baseline & 0.766 & 76.87\scriptsize{$\pm$0.24} & \multirow{2}{*}{1.9$\times$} \\
L1 ours (sparse) & 0.877 & 76.90\scriptsize{$\pm$0.22} & \\
\hline
\multicolumn{4}{l}{\textbf{VGG-16 on CIFAR-100}} \\
L1 baseline & 0.269 & 66.94\scriptsize{$\pm$0.24} & \multirow{2}{*}{2.4$\times$} \\
L1 ours (sparse) & 0.697 & 67.06\scriptsize{$\pm$0.62} & \\
\cdashline{1-4}
\multicolumn{4}{l}{\textbf{VGG-16 on CIFAR-10}} \\
L1 baseline & 0.808 & 91.29\scriptsize{$\pm$0.16} & \multirow{2}{*}{2.6$\times$} \\
L1 ours (sparse) & 0.926 & 91.38\scriptsize{$\pm$0.05} & \\
\hline
\multicolumn{4}{l}{\textbf{VGG-11 on SVHN}} \\
L1 baseline & 0.519 & 94.68\scriptsize{$\pm$0.08} & \multirow{2}{*}{3.1$\times$} \\
L1 ours (sparse) & 0.845 & 94.71\scriptsize{$\pm$0.01} & \\
\hline
\multicolumn{4}{l}{\textbf{VGG-16 on SVHN}} \\
L1 baseline & 0.450 & 95.34\scriptsize{$\pm$0.11} & \multirow{2}{*}{2.7$\times$} \\
L1 ours (sparse) & 0.795 & 95.38\scriptsize{$\pm$0.11} & \\
\hline
\multicolumn{4}{l}{\textbf{VGG-19 on SVHN}} \\
L1 baseline & 0.122 & 95.37\scriptsize{$\pm$0.11} & \multirow{2}{*}{9.9$\times$} \\
L1 ours (sparse) & 0.911 & 95.41\scriptsize{$\pm$0.07} & \\
\hline
\end{tabular}
\end{center}
\caption{ Compression rate, \textit{i.e.}, how much the non-zero-valued parameters is reduced, obtained by our sparse models.
}
\label{tab:sparsity}
\normalsize
\end{table}

L1 regularization naturally compresses neural networks by setting a portion of parameters to zero while it can even improve generalization with the simplified solutions.
In order to see how much sparsity our method can obtain while keeping the baseline's best accuracy, we choose a sparse model whose accuracy is equal to or higher than that of L1 baseline.
Our proposed model's sparsity and compression rate over the baseline are shown in Table \ref{tab:sparsity}.
Our model, in general, needs only about 10$\sim$30\% of all parameters to perform as good as the baseline, which needs about 20$\sim$90\% of all parameters.
Our approach always (7 out of 7) obtains higher sparsity with compression rate up to 9.9$\times$ than baselines, meaning that our approach is promising for compressing neural networks.

\subsection{Empirical Validation of Our Hypothesis}
\begin{figure}[tbh] 
\centering
\begin{subfigure}{.5\columnwidth}
    \centering
    \includegraphics[width=0.9\linewidth]{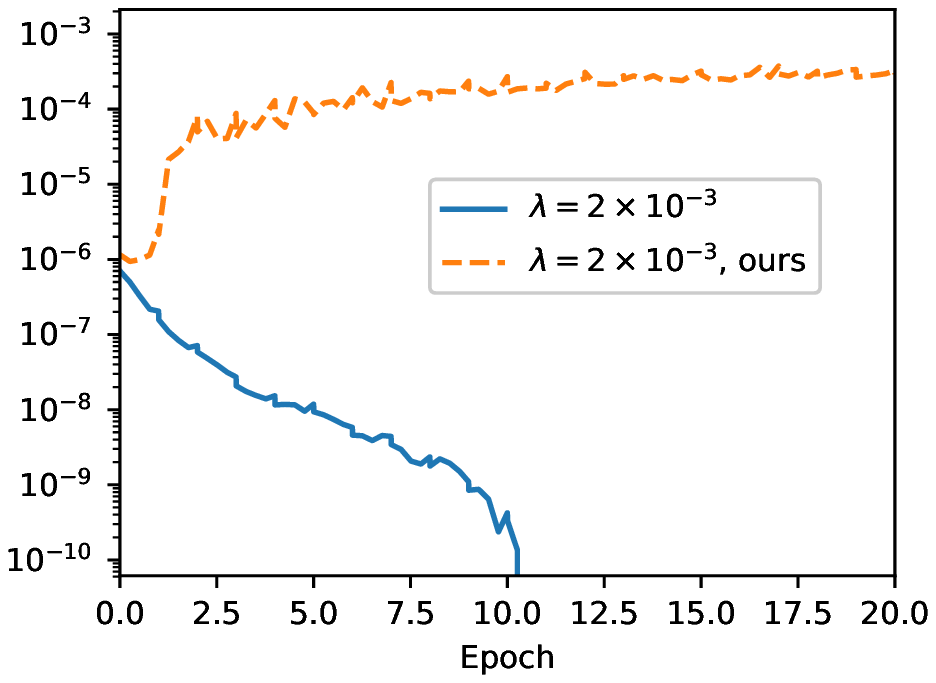}
    \caption{Avg. $|\nabla \mathcal{L}|$ (close-up)}
    \label{fig:grad_comp}
\end{subfigure}%
\begin{subfigure}{.5\columnwidth}
    \centering
    \includegraphics[width=0.9\linewidth]{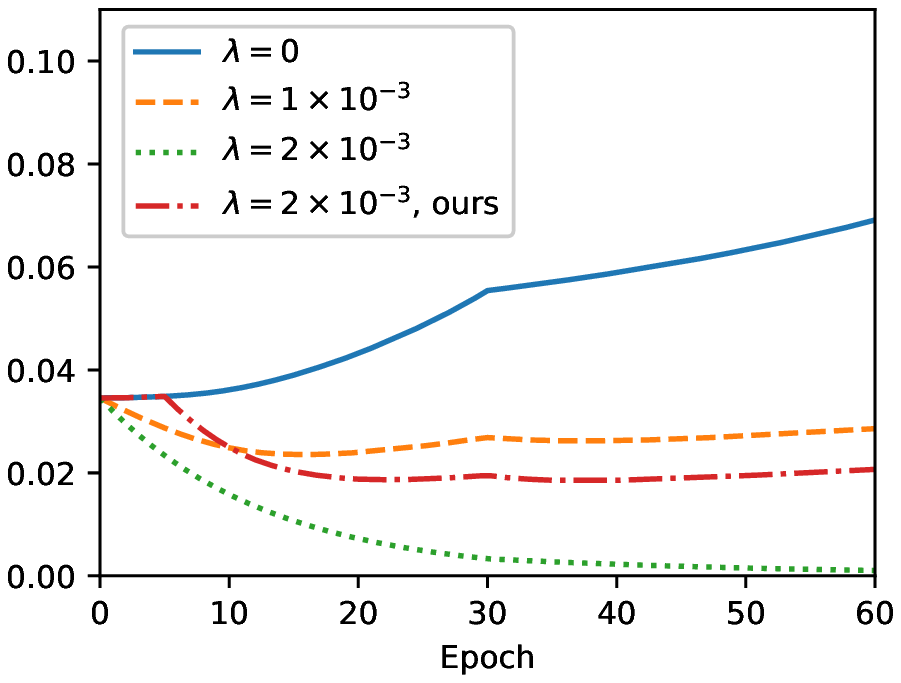}
    \caption{Avg. $|w|$}
    \label{fig:exp_weight_comp}
\end{subfigure}%
\caption{ Results by VGG-16 with L2 regularization on CIFAR-100. The results are from baseline unless it is labeled as ``ours''. }
\label{fig:comparison}
\end{figure}

We hypothesized that (\texttt{i}) if we skip strong regularization when the gradients are not coherent enough, the model will not fail to learn, and (\texttt{ii}) if the model does not suffer from continuous suppression in $|\nabla \mathcal{L}|$, then $|w|$ may not decrease.
It is shown in Figure \ref{fig:grad_comp} that our proposed model obtains great elevation instead of exponential decay in $|\nabla \mathcal{L}|$ unlike the baseline; this means that it indeed does not fail to learn.
In Figure \ref{fig:exp_weight_comp}, although the same strong regularization is enforced after a couple of epochs, the magnitude of weights in our model stops decreasing around epoch 20, while that in baseline (a green dotted line) keeps decreasing towards zero.
As there is no continuous suppression in $|\nabla \mathcal{L}|$ for our proposed model, the magnitudes of parameters indeed do not decrease after a certain point.
Comparing our approach with $\lambda=2\times 10^{-3}$ to baseline with $\lambda=1\times 10^{-3}$, we can also see that our approach with strong regularization indeed further simplifies the solution.

\section{Related Work}
The related work is partially covered in the introduction section, and we extend other related work here.
It has been shown that L2 regularization is important for training DNNs \cite{krizhevsky2012imagenet,deng2013new}.
Although there has been a new regularization method for DNNs such as dropout, L2 regularization has been shown to reduce the test error effectively when combined with dropout 
\cite{srivastava2014dropout}.
Meanwhile, L1 regularization has also been used often in order to obtain sparse solutions.
To reduce computation and power consumption, L1 regularization and its variations such as group sparsity regularization have been promising for deep neural networks 
\cite{wen2016learning,scardapane2017group,yoon2017combined}.
However, for both L1 and L2 regularization, the phenomenon that learning fails with strong regularization in DNNs has not been emphasized previously.
\cite{bergstra2012random} showed that tuning hyper-parameters such as L2 regularization strength can be effectively done through random search instead of grid search, but they did not study the phenomenon by strong regularization.
\cite{yosinski2015understanding} visualized activations to understand deep neural networks and showed that strong L2 regularization fails to learn.
However, it was still not shown how and why learning fails and how strong regularization can be achieved.
\cite{alvarez2016learning} applies a group sparsity regularizer only once per each epoch to compress DNNs, but the purpose is not to avoid failure in learning but to determine the number of necessary neurons in each layer.
To the best of our knowledge, there does not exist work that studies vanishing gradients and failure in learning caused by strong regularization.

\section{Discussion}\label{sec:discussion}

Our proposed method can be especially useful when strong regularization is desired.
For example, deep learning projects that cannot afford a huge labeled dataset can benefit from our method.
On the other hand, strong regularization may not be necessary in some other cases where the large labeled dataset is available or the networks do not contain many parameters.

Our work can be further extended in several ways.
Since our approach can achieve strong regularization, it will be interesting to see how our approach cooperates with the fixup initialization \cite{zhang2019fixup} that performs very well without normalization techniques but with proper initialization and regularization. 
In this work, we applied our approach to only L1 and L2 regularizers.
However, applying it to other regularizers such as group sparsity regularizers will be promising as they are often employed for DNNs to compress them.
Lastly, our proposed gradient-coherence algorithm is general, so one can apply it to other joint optimization problems where unfavorable gradients dominate in overall gradients.
All these directions are left for our future work.

{\small
\bibliographystyle{ieee}
\bibliography{mybib}
}

\end{document}